\documentclass[runningheads]{llncs}
\usepackage[margin=1in]{geometry}
\usepackage{graphicx}
\usepackage{hyperref}
\usepackage{amsmath}
\usepackage{algorithm}
\usepackage{algorithmic}
\begin{document}
\title{Deep neural network ensemble by data augmentation and bagging for skin lesion classification}
%
%
\author{Manik Goyal\inst{1} \and
Jagath C. Rajapakse\inst{2}}
%
%
\institute{Indian Institute of Technology, Varanasi, India\\
\email{ manik.goyal.cse15@iitbhu.ac.in}\\\and
Nanyang Technological University, Singapore,\\
\email{asjagath@ntu.edu.sg}}
\maketitle              
\begin{abstract}
This work summarizes our submission for the Task 3: Disease Classification of ISIC 2018 challenge in Skin Lesion Analysis Towards Melanoma Detection. We use a novel deep neural network (DNN) ensemble architecture introduced by us that can effectively classify skin lesions by using data-augmentation and bagging to address paucity of data and prevent over-fitting. 

The ensemble is composed of two DNN architectures: Inception-v4 and Inception-Resnet-v2. The DNN architectures are combined in to an ensemble by using a $1\times1$ convolution for fusion in a meta-learning layer.

\end{abstract}
\section{INTRODUCTION}
The Task 3 of ISIC 2018 challenge in Skin Lesion Analysis Towards Melanoma Detection \cite{isic18,isic17} is defined as to generate the binary classification corresponding to each of the 7 disease classes: melanoma, melanocytic nevus, basal cell carcinoma, actinic keratosis, benign keratosis, dermatofibroma and vascular lesion for each test image. The predicted responses are then scored using balanced normalized accuracy.

In this work, we propose a novel framework, data augmentation and bagging ensemble architecture (DABEA), that uses data augmentation and bagging in combination to generate multiple output vectors per model and then applies a $1\times1$ convolution layer as a meta-learner for combining different model outputs. Our main contributions includes using (i) an ensemble DNN models with data augmentation and bagging and (ii) a $1\times1$ convolution layer for meta-learning of models.

\section{Methodology}

\begin{figure*}[htbp]
\begin{center}
{\includegraphics[width=\textwidth]{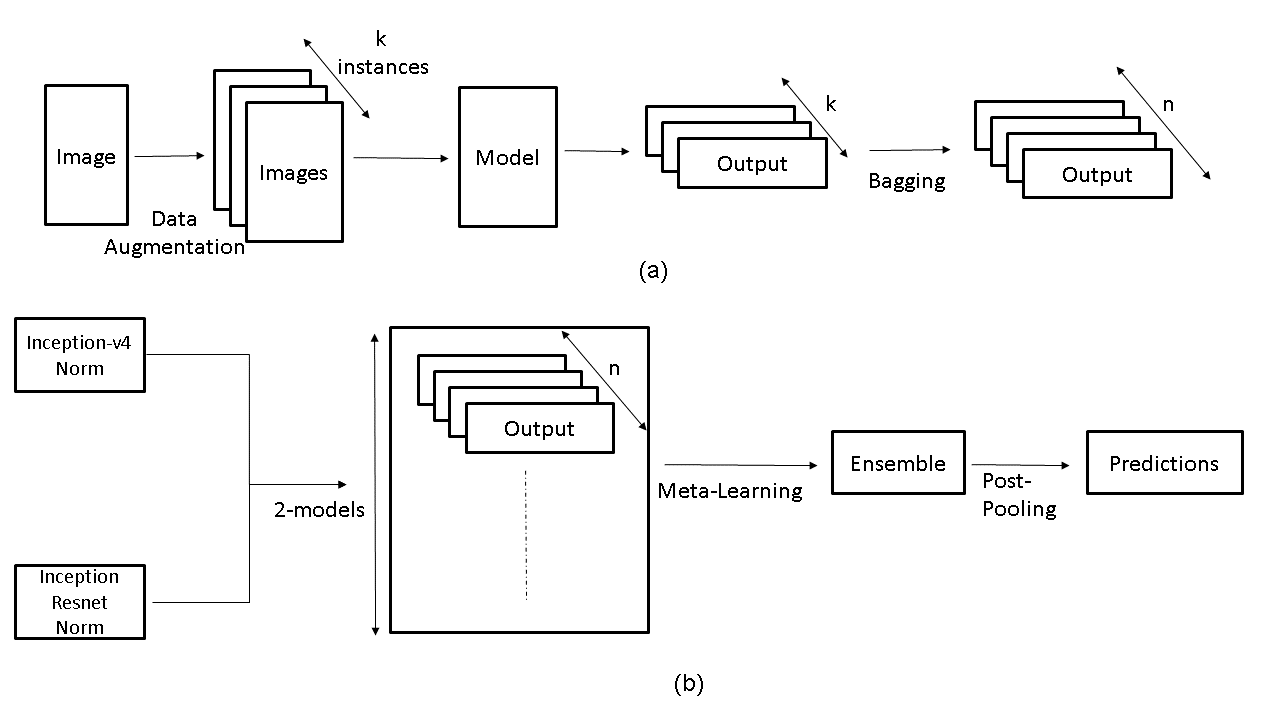}}
\end{center}
   \caption{Illustration of DABEA: (a) data augmentation and bagging of outputs for a single CNN model, and (b) concatenation of outputs of CNN models as inputs for the meta-learning fusion.}
\label{fig:ensemble} 
\end{figure*}

\subsection{Deep neural network models}
We selected two of the top performing CNN architectures namely Inception-v4 and Inception-Resnet-v2 \cite{inceptionv4} as our base training models because of their high performances accuracies on IMAGENET \cite{ILSVRC15} challenge. All these models are variants of Inception model that introduced the concept of filters with multiple sizes at the same level of CNN. Inception-v4 (Iv4) simplified the inception architecture by further factorization and introduced reduction blocks while Inception-Resnet-v2 (IRv2) combined residual blocks as in \cite{resnet} to deepen the model while having computational cost similar to Inception-v4.

\subsection{Data augmentation and bagging ensemble architecture (DABEA)}
Given a training set $({\bf X},{\bf Y})$ where ${\bf X}$ denotes the input data matrix consisting of $N$ input images of $d$ input dimensions and $\bf Y$ denotes the matrix consisting of corresponding labels of seven skin lesion classes. We build and train an ensemble of CNN architecture, or DABEA, for two-class classification of skin lesions. The DABEA combines two CNN models: (i) Inception-v4 (Iv4), and (ii) Inception-Resnet-v2 (IRv2). Parameters of both the models are first learnt by training on the IMAGENET data \cite{ILSVRC15} and then on ISIC 2018 skin lesion classification dataset train split \cite{isic18}.

First, the images are normalized by subtracting the per-image pixel mean values and obtained the normalized data ${\bf \widetilde{X}}$:
\begin{equation}
{\bf \widetilde{X}} = normalize(\bf{X})
\label{equ_norm}
\end{equation}
Normalization is used to remove any bias present in the data \cite{recod,pre-seg,Matsunaga}. Second, 
data augmentation is performed on the normalized images by cropping, random brightness and saturation changes, and flipping. The augmented dataset ${\bf X}_{a}$ increases the number of images available for training:
\begin{equation}
{\bf X}_{a} = augment({\bf \widetilde{X}})
\label{equ_aug}
\end{equation}
Let the original dataset is augmented by $k$ times. Training images are then fed to Inception-v4 and Inception-Resnet-v2 networks in cascade, so the ensemble consists of models $m \in \{ {\rm Iv4}, {\rm IRv2} \}$. Let ${\bf Y}_{m}$, $f_m$, and $\theta_m$ denote the output, transfer function, and parameters of the model $m$, respectively. The output ${\bf Y}_{m}$ is given by
\begin{equation}
{\bf Y}_{m} = f_{m}(\theta_m, {\bf X}_{a})
\label{equ_out}
\end{equation}
where $m \in\{ \rm{Iv4}, \rm{IRv2}\}$ and ${\bf Y}_{m} \in {\rm I\!R}^{Nk\times 7}$. 

Bagging is then performed by randomly selecting $n$ output vectors from each model. Let ${\bf \widetilde{Y}}_m$ denote the bagged feature output from CNN model $m$:
\begin{equation}
\widetilde{\bf Y}_m = randomselect({\bf Y}_{m})
\label{equ_bagg}
\end{equation}
Finally, the bagged data from two models are combined into one feature output ${\bf {Y}}$ as follows:
\begin{equation}
{\bf {Y}} = ({\bf \widetilde{Y}}_{m},m \in \{ {\rm Iv4}, {\rm IRv2} \} )
\label{equ_comb}
\end{equation}

\subsection{Meta-Learning Output Layer}
Ensemble meta learning is done  by providing ${\bf {Y}} \in {\rm I\!R}^{N\times 7\times n\times 2}$ feature vector to a $1\times 1$ convolution layer. The $1\times1$ convolutional layer combines the two input channels from the two CNN architectures and fused them by pooling to produced output ${\bf O}$
\begin{equation}
{\bf O} = pooling({\bf Y})
\label{equ_pool}
\end{equation}
Forward propagation of activation in the DABEA architecture is illustrated in Algorithm 1.  

\begin{algorithm}
\caption{Forward propagation of DABEA}
\begin{algorithmic}
\label{algorithm:1}
\STATE Given input images ${\bf X}$
\STATE $\widetilde{\bf X} = normalize({\bf X}$)
\STATE ${\bf X}_a = augment(\widetilde{\bf X}$)
\FOR{$m \in \{\rm Iv4, IRev2\}$}
    \STATE ${\bf Y}_m = f_{m}(\theta_m, {\bf X}_{a})$
    \STATE  $\widetilde{\bf Y}_m = randomselect( {\bf Y}_{m})$
\ENDFOR
\STATE $ {\bf {Y}} = ({\bf \widetilde{Y}}_{m},m \in \{ {\rm Iv4}, {\rm IRv2} \} )$ 
\STATE ${\bf O} = pooling({\bf Y})$
\end{algorithmic}
\end{algorithm}

\section{Experiments and Results}
We split the ISIC 2018 \cite{isic18,isic17} training data into 90:10 to form the internal training and validation split. Two base models, namely, Iv4 and IRv2 are first trained on the training split with weights pre-trained on IMAGENET data \cite{ILSVRC15}. The ensemble $1\times1$ convolution is learnt using DABEA feature output over the internal validation split. All the base models and ensemble convolution are trained using cross entropy loss. For producing predictions for every official test and validation image the model DABEA is used over the ISIC 2018 test and validation data.

We experimented with both normalized and un-normalized images as input for training the base models. For normalization we used the per-image normalization as suggested by \cite{recod}.

For training the Inception-v4 and Inception-Resnet-v2 models, we used Adam optimizer with a initial learning rate of 0.01 which decays over two epochs with an exponential rate of 0.94. All the normalized models were trained for 20000 epochs while un-normalized are trained for 40000 epochs with a dropout probability of 0.2. While training the ensemble part, we randomly bag $n=100$ output-vectors, each were produced by  $k=10$ augmented inputs.

We use a $1\times1$ convolution fusion layer\cite{1CNN} for fusing the outputs in the proposed ensemble CNN model ; The single $1\times1$ convolutional fusion layer is optimized using Adam optimizer with a constant learning rate of $1e^4$ and is trained for 100 epochs. The learnt weights are then used to obtain predictions on the official ISIC 2018 test and validation data. The outputs produced by the 100 different combinations are then clubbed by using a post-pooling technique. We experimented with 3 different pooling techniques: max-pooling, avg-pooling, and extreme-probability pooling. And finally fixed avg-pooling for test submission.

For DABEA ensemble we used three different base model sets: (i)Un-Normalized Iv4 and IRv2, (ii)Normalized Iv4 and IRv2, (iii)Both Norm. and Un-Norm Iv4 and IRv2. The performance measures for the three different sets over the ISIC 2018 validation data is given in the ~\autoref{table-ensemble}.

\begin{table}[]
\centering
\caption{Comparison of Balanced Accuracy values for $1\times1$ conv. ensemble using different base model sets over the ISIC 2018 official validation data\cite{isic18,isic17}}
\label{table-ensemble}
\begin{tabular}{|l|c|c|c|c|}
\hline
Model & Balanced Accuracy \\ \hline
$1\times1$ CNN Ensemble (Un-Norm.) & 0.729 \\ \hline
$1\times1$ CNN Ensemble (Norm.)& \textbf{0.837} \\ \hline
$1\times1$ CNN Ensemble (Norm. + Un-Norm.)& 0.732 \\ \hline
\end{tabular}
\end{table}

\end{document}